\documentclass{article}

\usepackage{PRIMEarxiv}

\usepackage[utf8]{inputenc} 
\usepackage[T1]{fontenc}    
\usepackage{hyperref}       
\usepackage{url}            
\usepackage{booktabs}       
\usepackage{amsfonts}       
\usepackage{nicefrac}       
\usepackage{microtype}      
\usepackage{lipsum}
\usepackage{fancyhdr}       
\usepackage{graphicx}       
\graphicspath{{media/}}     

\usepackage{natbib}
\usepackage{algorithm}
\usepackage{algorithmic}

\usepackage{amsfonts}       
\usepackage{amsmath}
\usepackage{amsthm}

\usepackage{booktabs}
\usepackage{multirow}

\usepackage{color}
\usepackage[none]{hyphenat}

\usepackage{authblk}

\title{Bandit-supported care planning for older people with complex health and care needs
}

\author[1,2]{\bf Gi-Soo Kim~}
\author[3]{\bf Young Suh Hong~}
\author[4]{\bf Tae Hoon Lee~}
\author[5,6]{\bf Myunghee Cho Paik~}
\author[4,7,8,\thanks{Corresponding author: Hongsoo Kim, \texttt{hk65@snu.ac.kr}}]{\bf Hongsoo Kim}
\affil[1]{Department of Industrial Engineering, UNIST, Ulsan, South Korea}
\affil[2]{Artificial Intelligence Graduate School, UNIST, Ulsan, South Korea}
\affil[3]{ Master of Health Informatics, School of Information, 
  University of Michigan-Ann Arbor, MI 48109, USA}
 \affil[4]{Department of Public Health Sciences, Graduate School of Public Health, Seoul National University, Seoul, South Korea}
 \affil[5]{Department of Statistics, Seoul National University, Seoul, South Korea}
 \affil[6]{Shepherd23 Inc., Seoul, South Korea}
 \affil[7]{Institute of Health and Environment, Graduate School of Public Health, Seoul National University, Seoul, South Korea}
 \affil[8]{AI Institute, Seoul National University, Seoul, South Korea}

\begin{document}
\maketitle

\begin{abstract}
Long-term care service for old people is in great demand in most of the aging societies. The number of nursing homes residents is increasing while the number of care providers is limited. Due to the care worker shortage, care to vulnerable older residents cannot be fully tailored to the unique needs and preference of each individual. This may bring negative impacts on health outcomes and quality of life among institutionalized older people. To improve care quality through personalized care planning and delivery with limited care workforce, we propose a new care planning model assisted by artificial intelligence. We apply bandit algorithms which optimize the clinical decision for care planning by adapting to the sequential feedback from the past decisions. We evaluate the proposed model on empirical data acquired from the Systems for Person-centered Elder Care (SPEC) study, a ICT-enhanced care management program.
\end{abstract}


\section{Introduction}

The world’s population is aging rapidly. In 2050, one fifth of the population will be over 60 years and about 80\% of the older people will be residing in Low- and Middle-Income Countries (LMICs) \citep{WHO2}. As the longevity is extended, long-term care needs of older people also increase dramatically. According to the recent OECD projection on the size of care workers needed in 2030 \citep{/content/publication/92c0ef68-en}, member countries anticipate severe shortage in number of care workers; in particular, South Korea experiencing the most rapid population aging needs about 140\% increase of long-term care workers assuming the care demands would be the same in 2030 as in 2016. In fact, shortage in professional care workers and lack of competency of personal care workers have impeded improvement in care quality in long-term care system in Korea \citep{KimKwon21}. It will be a real challenge to meet the rapidly increasing needs of skilled care workers in Korea as well as other developed countries. For LMICs, long-term care is still highly dependent on family members often without relevant training and formal support. Given the big global challenge in caring vulnerable older people in aging societies, policy-makers and business sectors explore the potential of innovation, but empirical evidence is sparse so far. In this paper, we focus on the feasibility and applicability of artificial intelligence-assisted health care in nursing homes.
The care provision in nursing homes can be framed as an online sequential decision problem. Once a resident is admitted to a nursing home, the care provider decides which interventions to apply. Health outcomes of a patient will be affected by the interventions (s)he received. Care provider learns from these feedbacks and reinforces his(her) decision for the next admitted resident. Bandit algorithms \citep{Li10, Chu11, Agrawal13} address the online decision problem. A learning agent faces a set of available actions (or arms) sequentially. The agent can execute only one action at a time, and receives a corresponding feedback (or reward). The expected value of the reward depends on the action executed and is unknown to the agent. The agent should learn the unknown values adaptively from the cumulated feedbacks. The goal of the agent is to reinforce the arm selection strategy sequentially so as to maximize the cumulative sum of rewards. We specifically focus on the contextual bandit problem, where side information called context is given on the reward and can be used when selecting the action. Popular contextual bandit algorithms include the Upper-Confidence Bound algorithm (LinUCB; \cite{Li10, Chu11})  and the Thompson Sampling algorithm for linear payoffs (LinTS; \cite{Agrawal13}) .
In this work, we propose a new care planning model assisted by LinUCB and LinTS. To evaluate the efficacy of care of the proposed framework, we analyze the data acquired from the Systems for Person-centered Elder Care (SPEC) study \citep{Kim17, Kim21}. The SPEC study is a randomized controlled trial to examine the effectiveness of ICT-enhanced integrated care management on quality of care for older residents in nursing homes of South Korea. In the study, we analyze data of 278 residents from 10 nursing homes who participated in the SPEC program. The care plan for each resident was developed by one of 20 nurses (SPEC care managers) in the nursing homes through care planning meetings with his(her) multi-disciplinary team. The number of residents who were assigned to a care manager for care planning varied from 1 to 31. The feedback of reward is preventing the loss of activities of daily living (ADL), a very critical goal of nursing home care. The care to prevent ADL loss was provided as the combination of 20 different intervention items, which include assessing new acute health problems, risk factor management, referral to therapy specialists, setting the care priorities, etc. Among the 20 interventions, the minimum number of interventions applied to a resident was 1 and the maximum was 16. The care provider applied a combination of interventions based on the observed characteristics of the older resident such as age, sex, length of stay, cognition, and baseline ADL.
We apply the bandit algorithms on the data and compare their performance with that of the nurses (care managers). 

\section{Algorithms}

\subsection{Settings}

We assume that residents are admitted to nursing homes sequentially and at each time $t$, a patient with characteristics $X_t$ is assigned to the agent. The covariate $X_t$ includes the age, gender, length of stay, cognition level, baseline ADL score. {\color{black}Based on available information in $X_t$, {the agent chooses an action, i.e., a combination of interventions, and applies it to the patient. We restrict the agent to choose the same number of interventions as was originally applied to the patient by the care provider. Also, we restrict the type of combinations to the ones that were realized in the data. For example, if a patient originally received 3 interventions, the action set that the bandit agent can choose from is a subset of the set of $\binom{20}{3}$ possible combinations. Hence, the number of available actions changes with time $t$. We denote the number of available actions at each time $t$ as $N_t$ }}
and we denote the index of the chosen action as $a(t)$. We also denote the reward of the $i$-th action for the resident at time $t$ as $r_{i,t}$. Among $r_{1,t}, r_{2,t}, \cdots, r_{N_t,t}$, the agent observes only one reward, $r_{a(t),t}$. This is the binary value indicating the prevention of ADL loss (1: prevention/ 0: no prevention).

\subsection{LinUCB and LinTS}

LinUCB and LinTS assume that the reward follows a linear model with sub-gaussian errors as follows. 
$$r_{i,t}=b_{i,t}^T\mu+\varepsilon_{i,t}.$$
The vector $b_{i,t}$ denotes the context variable of the $i$-th action at time $t$. We assume that $b_{i,t}$ is $d$-dimensional. We will specify the components of $b_{i,t}$ in our study in later sections. The coefficient vector $\mu$ is an unknown $d$-dimensional vector and should be learned by the learning agent. The scalar error $\varepsilon_{i,t}$ is assumed to be independent over $i$ and $t$ and to be sub-Gaussian with unknown variance. LinUCB is a bandit algorithm for linear payoffs proposed by \cite{Li10}. At each time $t$ , the algorithm updates the high-probability upper confidence bound (UCB) of the expected reward of each action and selects the action that has the highest value. The UCB of the $i$-th action at time $t$ has the following form, 
$$UCB_i(t)=b_{i,t}^T\hat{\mu}(t)+\alpha\sqrt{b_{i,t}^TB(t)^{-1}b_{i,t}},$$
for some positive constant $\alpha$ where 
$$B(t)=\lambda I_d+\sum_{\tau=1}^{t-1}b_{a(\tau),\tau}b_{a(\tau),\tau}^T$$
for some positive constant $\lambda$ and $d\times d$ identity matrix $I_d$ and $\hat{\mu}(t)$ is the Ridge estimator of $\mu$ fitted on the context-reward pair of the selected actions up to time $t-1$, i.e., 
$$\hat{\mu}(t)=B(t)^{-1}\sum_{\tau=1}^{t-1}b_{a(\tau),\tau}r_{a(\tau),\tau}.$$
This UCB is the sum of the current estimate of the reward and its uncertainty. Therefore, selecting the action with highest UCB means to select an action which has either a high estimate (“exploitation”) or high uncertainty (“exploration”). Exploration helps reducing the uncertainty related to that arm and it helps to make better choice in the future. Balancing between exploitation and exploration is crucial in achieving a high total reward.  In LinUCB, this balance is controlled by the constant $\alpha$. In \cite{Abbasi11}, authors derived the value of $\alpha$  which makes the UCB a tight bound that holds with high probability and consequently, guarantees of high cumulative rewards. This value is determined by various model parameters including the variance of the error and hence, is unknown to the learning agent. In practice, we need to tune this parameter by grid search. Bigger this value, more the algorithm will concentrate on exploration rather than exploitation. Exploration helps to learn the unknown rewards but does not help in maximizing the payoff. On the other hand, exploitation concentrates on maximizing the reward but may be misguided if the learning is not sufficient. Therefore, we should carefully tune this parameter in order to achieve our ultimate goal, maximizing the cumulative sum of rewards.

LinTS is similar to LinUCB, except that it uses Bayesian ideas to control the exploitation-exploration tradeoff. LinTS makes an additional assumption that the noise of the reward is normally distributed and selects the action randomly according to the posterior probability that it is the best arm, i.e., the arm that has the highest expected reward. Under the assumption that the prior distribution of $\mu$  is $\mathcal{N}(0_d,v^2I_d)$ where $0_d$ is a zero-valued $d$-dimensional vector and the error is normal $\mathcal{N}(0,v^2)$ for some positive constant $v$, it can be shown that the procedure is equivalent to sampling $\tilde{\mu}(t)$  from a multivariate normal distribution $\mathcal{N}(\hat{\mu}(t),v^2B(t)^{-1})$  and then selecting the arm with highest value of $b_{i,t}^T\tilde{\mu}(t)$. Note that when $\tilde{\mu}(t)\sim \mathcal{N}(\hat{\mu}(t),v^2B(t)^{-1})$, then $b_{i,t}^T\tilde{\mu}(t) \sim \mathcal{N}(b_{i,t}^T\hat{\mu}(t),v^2b_{i,t}^TB(t)^{-1}b_{i,t}).$ Here, we can observe that the standard deviation of the distribution is equivalent to the uncertainty term in LinUCB. The value $v$  is analogous to the value $\alpha$  in the LinUCB and is also a hyper parameter to tune.

\section{Methods of Experiments}

\subsection{Preprocessing of the data}\label{preprocess}

{\subsubsection{ Inherent incompleteness of real data }}

The data constitutes of the covariates, intervention types, and the indicator of ADL loss prevention of 278 patients in 10 nursing homes, who participated in the six-month SPEC intervention study from April 2015 to December 2016 (a total of 21 months trial). The covariates $X_t$  of the $t$-th patient include the age, gender, cognition, ADL, and additional variables (hearing, depression, pain, comorbidities, number of comorbidities) of older residents. A resident could receive a subset of 20 different interventions. We made a binary indicator $I_k(t)$  for the $k$-th intervention to indicate whether or not it was provided to the $t$-th patient. {\color{black}Among the residents, 13.3\% experienced ADL loss.}

A challenge in applying the bandit algorithm to this dataset is that the reward from every action is not  observed. For example in our dataset, for each patient incoming at time $t$, only one action that a nurse/care team selected is realized and the rewards of the other $N_t-1$  actions are unobserved. We denote the missing rewards as counterfactuals, as they are the outcomes that would have been observed if a different action had been applied. That is, for each $t$, only $r_{\tilde{a}(t),t}$ is observed among $r_{1,t}, r_{2,t}, \cdots, r_{N_t,t}$  where $\tilde{a}(t)$  denotes the realized action in the dataset. Therefore, if the action choice of the bandit algorithm does not match the one that is realized, i.e., if $\tilde{a}(t)\neq a(t)$, we cannot provide any feedback to the algorithm. 

{\subsubsection{ Synthesis of `full’ data }}

To circumvent this problem, we impute the missing rewards for each patient using a machine learning model which fits the data well. {\color{black}To find the best model, we fitted Support Vector Machine (SVM; \cite{cortes1995support}), Random Forest (RF; \cite{ho1995random}), and Light Gradient Boosting Machine (LGBM; \cite{ke2017lightgbm}) to the data using the observed binary reward $r_{\tilde{a}(t),t}$ as outcome and the $X_t$ and intervention indicators $I_1(t), I_2(t), \cdots, I_{20}(t)$ as covariates. For the SVM, we used either the linear or radial basis function kernel kernel. For RF, we tried different combinations of number of trees, maximum number of features used to split the nodes and the maximum depth of the trees. For LGBM, we tried different combinations of number of trees, number of leaves, maximum depth, learning rate, L1 penalization and L2 penalization parameters. Also for all the models, we implemented two different versions, {one that upweights the samples in the minority class with $r_{\tilde{a}(t),t}=0$ and the other which uses equal weights. }
We compared the Area Under the ROC Curve (AUC) using a 5 fold cross-validation samples. The LGBM with 50 trees, 30 leaves, maximum depth 20, learning rate 0.1 using upweighting and with both L1 and L2 regularizations achieved the highest AUC, 0.65284. After selecting the model, we refitted the model on the whole data using the same hyperparameters. 
{We verified that by setting the classification threshold to 0.26, our model achieves training accuracy 0.996. Thus, we consider our model as one possible scenario that might have generated the observed rewards. From the next section, we treat this model as the true model that should be learned by bandit algorithms. We emphasize that the goal of this process is to synthesize a ‘full’ real data including  the rewards for the arms that the real setting did not choose. }
}
 
\subsection{Application of bandit algorithms}

We applied both LinUCB and LinTS to the fully imputed data. We randomly sampled one patient at a time (allowing repetition) to simulate the situation where patients are admitted to a nursing home sequentially. Each time a resident comes in, LinUCB and LinTS select a subset of 20 interventions respectively, receive the corresponding rewards (provided by the LGBM in Section \ref{preprocess}), and update the estimates $\hat{\mu}(t)$ and $B(t)$, which are used in the action selection at the next step. Then the next older resident (randomly sampled) comes in and the process is repeated. 

To describe the procedure in detail, we introduce new notations. We let $I^i=[I^i_1, I^i_2, \cdots, I^i_{20}]^T$  denote the set of the 20 binary indicators of the interventions for the $i$-th action, i.e., the $i$–th combination among the $N_t$  possible combinations. We let $\tilde{X}_t$  be the subset of the full covariates $X_t$ which includes only the age, gender, length of stay, cognition, and baseline ADL of a resident. We assume that only $\tilde{X}_t$  is observed and the remaining variables in $X_t$  are not available to the learning agent so cannot be used for action selection. Note that in Section \ref{preprocess}, we used the full covariates $X_t$  to fit the LBGM. This is because the LGBM was used to impute the missing rewards, not to make an action decision.

\begin{figure}[tp]
 \begin{center}
 \includegraphics[width=0.9\textwidth]{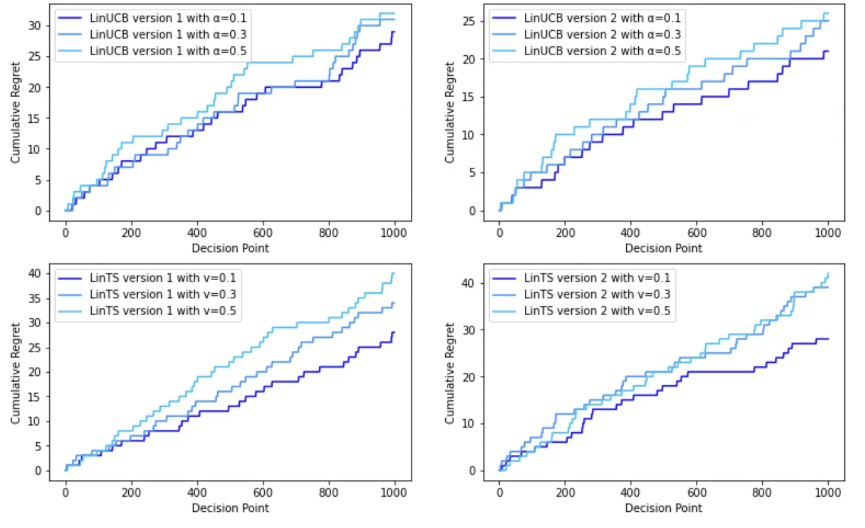}
  \caption{Median of cumulative regret of LinUCB and LinTS over 5 experiments for different $\alpha$'s and $v$'s}\label{fig:UCB_TS}
 \end{center}
 \end{figure}

 \begin{figure*}[thp!]
 \begin{center}
 \includegraphics[width=0.9\textwidth]{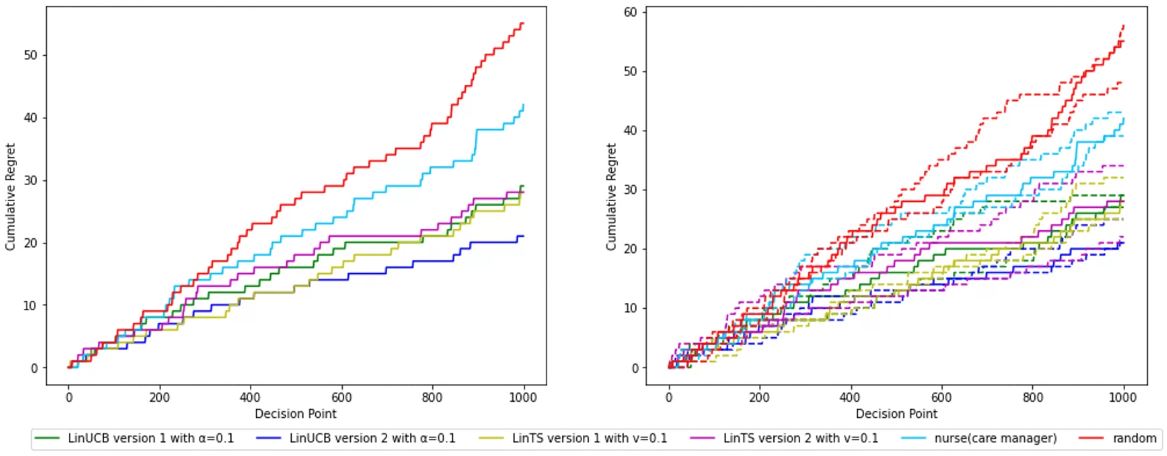}
 \caption{(Left) Median(solid line) of cumulative regret of each algorithm over 5 experiments (Right) Median(solid line) along with 25\% and 75\% percentiles of cumulative regret of each algorithm over 5 experiments}\label{fig:all}
 \end{center}
 \end{figure*}

We ran two versions of each algorithm depending on the linear model that the algorithm assumes. The first version assumes that the reward model is
$$r_{i,t}=\tilde{X}_t^T\mu_1+I^{iT}\mu_2+\varepsilon_{i,t}$$ 
and second version assumes
$$r_{i,t}=\tilde{X}_t^T\mu_1+I^{iT}\mu_2+(\tilde{X}_t \circ I^i)^T\mu_3+\varepsilon_{i,t},$$
where $\tilde{X}_t \circ I^i$  denotes the set of all possible interaction terms between $\tilde{X}_t$ and $I^i$. In the first model, $b_{i,t}=\left[\tilde{X}_t^T, I^{iT}\right]^T$ and $\mu=\left[\mu_1^T, \mu_2^T\right]^T$. In the second model, $b_{i,t}=\left[\tilde{X}_t^T, I^{iT}, (\tilde{X}_t \circ I^i)^T\right]^T$ and $\mu=\left[\mu_1^T, \mu_2^T, \mu_3^T\right]^T$.

In the first model, the best action is identical for all patients irrespective of the patient characteristics. The effect of the $\tilde{X}_t$  on the reward is only additive to the intervention indicators. On the other hand, in the second model, the best action is allowed to vary according to the patient characteristics. This allows personalized care by the bandit algorithm.

The second model includes the first model as a special case, and hence is more flexible. However, the second model has more parameters to learn (the coefficients of the interaction terms), so the optimization can be much slower than the first model. In the bandit setting, fast learning is one of the desired properties, especially when the time horizon is small. We compare the performance of the two algorithms, and compare it with the performance of the nurses (care manager).

We tuned the hyperparameters – $\alpha$ in LinUCB and $v$  in LinTS – by grid search. We tried 3 different candidate values, 0.1, 0.3, and 0.5, and chose the value that resulted in highest cumulative reward.




\section{Results}



We observe the regret of each algorithm which is defined as $regret(t)=r_{a^*(t),t}-r_{a(t),t}$  at each time $t$ , where $a^*(t)$ is the optimal action, i.e., the action which achieves the highest reward. Figure \ref{fig:UCB_TS} shows the cumulative regret of LinUCB and LinTS for each version of the models according to 3 different values of the hyper parameters. For each of the 4 cases, we chose the best value of the hyper parameter and used them to generate Figure \ref{fig:all}.

Figure \ref{fig:all} depicts the cumulative regret of the bandit algorithms for each version of the model along with a purely random algorithm and the nurses’s decision. The care manager's decision corresponds to the realized action in the dataset, i.e., $\tilde{a}(t)$.  {\color{black}Interestingly, we can observe that up to the 21st patient, the nurses’ performance is the best, but the bandit algorithms start to outperform as soon as they learn the reward model. In this experiment, LinUCB appears to perform better than LinTS. Among the two versions of the reward model, the second one which assumes interaction effects between the intervention and the patient characteristics outperforms the other one. }Further study on the performance of the two algorithms may be required.

\section{Discussion}

This study supports the idea that bandit algorithms may help shortage of human experts by assisting tailored care planning for each older person with complex care needs.

The algorithm learns the feedback mechanism systematically. At the beginning of the deployment, the model has high uncertainty and the algorithm concentrates on exploring the large action space to collect data that help reduce the uncertainty. As data accumulate, the learned model converges to the true model and the algorithm concentrates on exploitation, maximizing the rewards. This systematic care planning model can help human care managers to decrease unnecessary variation in their performance.

While the experiment results in this paper are promising, discussions on safety should be intensively conducted before deploying bandit algorithms in health care settings. Bandit algorithms require sufficient exploration of the action space to accelerate the learning process.  Consequently, the algorithm suffers higher regret than skilled care providers at the beginning of the deployment. While any intervention considered in this study is unlikely to have bad effect on the residents, so that applying a different action from the optimal one would hardly result in dangerous outcomes, there could be other health care settings where such unrestricted exploration could cause adverse events. Future work on bandit algorithms with safety constraints or blending the knowledge of the skilled care providers to the bandit algorithm would be beneficial.


Further studies are guaranteed to advance AI-assisted care decision-support under conditions with limited human and financial resources. This study has focused on a single critical outcome, prevention of late loss of ADL; but further studies can apply bandit care models for other important outcomes related to the well-being of frail older people, such as cognitive decline, delirium, and/or behavioral problems. Future studies can also develop prediction models for multiple outcomes, which are more likely in real-world care settings, where older people have two or more care needs and nurses make care decisions for complex care needs simultaneously. 

\bibliographystyle{plainnat}  
\bibliography{biblio}

\end{document}